\documentclass[11pt]{article}

\usepackage[T1]{fontenc}
\usepackage[utf8]{inputenc}
\usepackage{lmodern}

\usepackage[a4paper,margin=1in]{geometry}
\usepackage{setspace}
\usepackage{amsmath,amssymb,amsfonts,amsthm}
\usepackage{mathtools}
\usepackage{bm}
\usepackage{bbm}   % indicator function \mathbbm{1}

\usepackage{booktabs}
\usepackage{multirow}
\usepackage{array}
\usepackage{tabularx}
\usepackage{threeparttable}
\usepackage{siunitx}
\usepackage{enumitem}
\setlist[itemize]{noitemsep,topsep=2pt,leftmargin=*}
\setlist[enumerate]{noitemsep,topsep=2pt,leftmargin=*}

\usepackage{graphicx}
\usepackage{caption}
\usepackage{subcaption}
\usepackage{float}
\usepackage{wrapfig}

\usepackage{tikz}
\usetikzlibrary{arrows.meta,positioning,calc,fit}
\usepackage{pgfplots}
\pgfplotsset{compat=1.18}

\usepackage[ruled,vlined,linesnumbered]{algorithm2e}
\SetKwComment{Comment}{$\triangleright$ }{}

\usepackage[dvipsnames]{xcolor}
\usepackage{xurl}
\usepackage{url}

\usepackage[
  colorlinks=true,
  linkcolor=MidnightBlue,
  citecolor=MidnightBlue,
  urlcolor=MidnightBlue
]{hyperref}

\theoremstyle{plain}
\newtheorem{theorem}{Theorem}[section]

\theoremstyle{definition}

\newtheorem{assumption}[theorem]{Assumption}

\theoremstyle{remark}

\begin{document}

\title{Greedy Is Enough: Sparse Action Discovery in Agentic LLMs}
\author{Angshul Majumdar}
\date{}
\maketitle

\begin{abstract}
Modern agentic systems operate in environments with extremely large action spaces, such as tool-augmented language models with thousands of available APIs or retrieval operations. Despite this scale, empirical evidence suggests that only a small subset of actions meaningfully influences performance in a given deployment. Motivated by this observation, we study a contextual linear reward model in which action relevance is governed by a structured sparsity assumption: only a small number of actions have nonzero effects across latent states.

We formulate action discovery as a block-sparse recovery problem and analyze a greedy algorithm inspired by Orthogonal Matching Pursuit. Under standard assumptions on incoherence, signal strength, and action coverage, we prove that the greedy procedure exactly recovers the relevant action set with high probability, using a number of samples that scales polynomially in the sparsity level and latent dimension, and only logarithmically in the total number of actions. We further provide estimation error guarantees for refitted parameters and show that the resulting decision rule is near-optimal for new latent states.

Complementing these results, we establish information-theoretic lower bounds demonstrating that sparsity and sufficient coverage are necessary for tractability. Together, our results identify sparse action discovery as a fundamental principle underlying large-action decision-making and provide a theoretical foundation for action pruning in agentic systems.
\end{abstract}

\section{Introduction}

Large language models (LLMs) are increasingly deployed as interactive decision-making systems that operate beyond passive text generation. Modern deployments allow LLMs to invoke external tools, query document repositories, call APIs, or trigger environment-altering actions during inference. As a result, the effective action space faced by such systems can easily grow to tens or hundreds of thousands of discrete actions. This phenomenon is now commonly referred to as \emph{agentic} behavior.

Despite the scale and practical importance of these systems, current agentic LLM architectures rely almost entirely on heuristic control mechanisms. Tool selection is typically handled via prompting, hand-designed execution graphs, or finite-state workflows that specify when and how tools may be invoked. While effective in practice, these approaches provide no formal guarantees on optimality, stability, or sample efficiency, and offer little insight into the fundamental computational challenges posed by large action spaces.

A key empirical observation motivates this work. Although the nominal action space available to an LLM agent may be massive, successful behavior in any fixed task distribution consistently relies on only a small subset of actions. Most tools or documents are never used, while a few are repeatedly invoked across similar contexts. This suggests that the primary difficulty in agentic decision-making is not linguistic reasoning, but rather \emph{control under extreme action dimensionality with latent sparsity}.

This observation naturally reframes tool-augmented LLMs \cite{yao2023react} as instances of a broader class of decision problems: sequential control with a large discrete action space, where optimal behavior admits a sparse representation. From this perspective, the central challenge is to identify, from data, the small subset of actions that are relevant in different contexts, and to do so with computational and sample complexity that scales favorably with the action dimension.

In this paper, we formalize this problem and study it from a theoretical standpoint. While the interaction between an agent and its environment is naturally modeled as a partially observable Markov decision process (POMDP), we deliberately isolate a single orthogonal difficulty: the discovery of relevant actions when the action space is extremely large. We therefore assume access to a fixed state representation, which may be interpreted as a belief state, intent embedding, or any sufficient statistic produced by an upstream inference mechanism. Our focus is on sparse action discovery conditioned on this representation.

We introduce a model in which action relevance depends on a low-dimensional latent state, and where, for any given state, only a small number of actions have non-negligible influence on the reward. Importantly, the set of relevant actions is allowed to vary across states, but does so in a structured manner governed by a shared latent representation. This yields a form of \emph{state-dependent sparsity}, which generalizes classical global sparsity assumptions and captures the behavior observed in tool-augmented LLM systems.

To address this setting, we propose a greedy action selection method inspired by Orthogonal Matching Pursuit (OMP). Rather than treating tool selection as a linguistic or architectural design problem, our approach formulates it as a sparse recovery problem over actions. The resulting algorithm incrementally identifies relevant actions by correlating observed rewards with contextual features, and constructs a compact action set that supports near-optimal decision-making.

Our contributions are summarized as follows:
\begin{itemize}
    \item We formalize agentic tool use as a sequential decision problem with massive action spaces and structured, state-dependent sparsity.
    \item We propose a greedy, OMP-style algorithm for discovering relevant actions conditioned on a latent state representation.
    \item We establish support recovery guarantees showing that the relevant action set can be identified with sample complexity scaling logarithmically in the number of available actions.
    \item We show that without explicit sparsity assumptions, any method must incur sample complexity linear in the action dimension, demonstrating the necessity of sparse control.
\end{itemize}

Our analysis does not depend on the internal structure of language models. LLMs serve only as a motivating example of systems that expose a previously underexplored regime of decision-making: control with extreme action dimensionality and latent sparsity. By isolating and analyzing this regime, we aim to provide a theoretical foundation for agentic behavior that extends well beyond language-based systems.

\subsection{Interpretation in Agentic LLM Systems}

Although the formulation studied in this paper is entirely model-agnostic, it is useful to briefly relate the abstract objects appearing in our analysis to their counterparts in contemporary agentic LLM systems. This interpretation is intended solely as a semantic bridge and plays no role in the technical development.

\paragraph{Latent state.}
The latent state \(z_t \in \mathbb{R}^d\) represents a low-dimensional summary of the agent’s information at time \(t\). In an agentic LLM system, \(z_t\) may correspond to an internal representation produced by the language model after processing the interaction history, including user inputs, previous responses, and any tool outputs. Examples include hidden states, intent embeddings, or belief-like summaries. Throughout the paper, \(z_t\) is treated as given; we make no assumptions about how it is computed or updated.

\paragraph{Actions.}
Actions \(a \in \mathcal{A}\) correspond to discrete external operations available to the agent. In the LLM setting, these include tool invocations, document retrieval queries, API calls, database accesses, or other environment-interacting commands. The action space \(\mathcal{A}\) may be extremely large, reflecting the scale of modern tool libraries or document collections.

\paragraph{Reward.}
The reward \(r_t\) models task-level feedback used to evaluate agent performance. In practice, this may represent task completion, correctness of an answer, user satisfaction, or any scalar signal derived from downstream evaluation. Our analysis does not require the reward to be immediate, deterministic, or explicitly engineered, and applies equally to delayed or noisy feedback.

\paragraph{State-dependent sparsity.}
The central modeling assumption of this paper is that, for any given latent state \(z_t\), only a small subset of actions in \(\mathcal{A}\) have non-negligible influence on the reward. In the context of agentic LLMs, this reflects the empirical observation that only a few tools or documents are relevant to a given user intent, while the vast majority are irrelevant. Importantly, the identity of these relevant actions may vary across states, but does so in a structured manner governed by the shared latent representation.

\paragraph{Greedy action discovery.}
The greedy action selection procedures studied in this paper correspond, in agentic LLM systems, to the incremental identification of useful tools based on empirical performance, rather than reliance on hand-designed workflows, prompts, or execution graphs. Our results characterize when such greedy selection is sufficient to recover the relevant action set and enable near-optimal behavior.

This interpretation allows the theoretical results developed in the subsequent sections to be read directly as statements about the feasibility and limitations of tool selection in agentic LLM systems, while preserving the generality of the underlying mathematical framework.

\section{Problem Formulation}

We consider a sequential decision problem with a large discrete action space and structured, state-dependent sparsity. This section introduces the formal model and assumptions that underlie the greedy action discovery methods studied in the remainder of the paper.

\subsection{Latent State and Actions}

Let \(z_t \in \mathbb{R}^d\) denote a latent state representation available to the agent at time \(t\), where \(d \ll M\). The latent state summarizes all information relevant for decision-making at time \(t\). Throughout this paper, \(z_t\) is treated as given and fixed; we do not model how it is generated or updated.

The agent selects an action \(a_t\) from a finite action set
\[
\mathcal{A} = \{1,2,\dots,M\},
\]
where \(M\) may be extremely large. We allow \(M\) to scale polynomially or exponentially with problem parameters, reflecting settings in which the agent has access to a large number of tools, documents, or external operations.

\subsection{Reward Model}

After selecting action \(a_t\), the agent observes a scalar reward \(r_t \in \mathbb{R}\). We assume the reward satisfies the linear model
\begin{equation}
r_t = \langle W^\star_{a_t}, z_t \rangle + \varepsilon_t,
\label{eq:reward_model}
\end{equation}
where:
\begin{itemize}
    \item \(W^\star \in \mathbb{R}^{M \times d}\) is an unknown parameter matrix,
    \item \(W^\star_{a}\) denotes the \(a\)-th row of \(W^\star\),
    \item \(\varepsilon_t\) is zero-mean noise.
\end{itemize}

The linear structure in \eqref{eq:reward_model} serves as a realizability assumption and allows us to focus on the statistical and computational challenges induced by the large action space. Extensions to generalized linear or weakly nonlinear reward models are discussed in Section~\ref{sec:discussion}.

\subsection{State-Dependent Sparsity}

The central structural assumption of this paper is that the reward depends on only a small subset of actions.

\begin{assumption}[Row-Sparse Action Influence]
\label{assump:sparsity}
There exists an unknown index set \(S^\star \subseteq \mathcal{A}\) with \(|S^\star| = k \ll M\) such that
\[
W^\star_a = 0 \quad \text{for all } a \notin S^\star.
\]
\end{assumption}

Under Assumption~\ref{assump:sparsity}, only actions in \(S^\star\) have nonzero influence on the reward for any latent state \(z_t\). Importantly, while the relevance of actions varies with \(z_t\) through the inner product in \eqref{eq:reward_model}, the set of potentially relevant actions is governed by a shared sparse structure.

This assumption captures a form of \emph{state-dependent sparsity}: for a given state \(z_t\), the reward may depend on different actions to different degrees, but the total number of actions that can ever influence the reward is small.

\subsection{Data Collection Model}

We assume access to a dataset
\[
\mathcal{D}_T = \{(z_t, a_t, r_t)\}_{t=1}^T,
\]
where the actions \(a_t\) may be selected adaptively or non-adaptively. Our theoretical guarantees do not depend on a specific exploration strategy, provided the resulting data satisfy mild coverage conditions stated in Section~\ref{sec:theory}.

For notational convenience, define the feature vector
\[
\psi(z_t, a_t) = e_{a_t} \otimes z_t \in \mathbb{R}^{Md},
\]
where \(e_a\) denotes the \(a\)-th standard basis vector in \(\mathbb{R}^M\). Under this representation, the reward model \eqref{eq:reward_model} can be written compactly as
\[
r_t = \langle \theta^\star, \psi(z_t, a_t) \rangle + \varepsilon_t,
\]
where \(\theta^\star = \mathrm{vec}(W^\star)\) is block-sparse with \(k\) nonzero blocks.

\subsection{Objective}

The goal is to identify the unknown support set \(S^\star\) from the data \(\mathcal{D}_T\), and thereby reduce the effective action space from size \(M\) to size \(k\). Once \(S^\star\) is identified, decision-making can be restricted to this reduced action set, enabling efficient control and planning.

In the following section, we propose a greedy, OMP-style algorithm for recovering \(S^\star\) with sample complexity scaling logarithmically in \(M\).

\section{Greedy Sparse Action Discovery}

In this section, we introduce a greedy algorithm for identifying the relevant action set \(S^\star\) from data. The algorithm is inspired by Orthogonal Matching Pursuit (OMP) \cite{tropp2007omp} and exploits the block-sparse structure \cite{eldar2010blocksparse} induced by the reward model in Section~2. We emphasize that the procedure is entirely generic and does not depend on any language-model-specific assumptions.

\subsection{Motivation}

When the action space is extremely large, direct optimization over all actions is computationally infeasible. However, under Assumption~\ref{assump:sparsity}, only a small subset of actions influences the reward. The primary goal is therefore to recover the support set \(S^\star\) rather than to estimate the full parameter matrix \(W^\star\).

From a practical perspective, this corresponds to identifying a small set of candidate actions that are potentially useful across the observed latent states. In agentic LLM systems, this mirrors the informal practice of narrowing down a large tool library to a small subset of relevant tools before performing any downstream reasoning or planning. Here, we formalize this intuition as a statistical recovery problem.

\subsection{Design Matrix and Block Structure}

Recall the feature representation
\[
\psi(z_t, a_t) = e_{a_t} \otimes z_t \in \mathbb{R}^{Md},
\]
and define the design matrix
\[
\Psi \in \mathbb{R}^{T \times Md},
\]
whose \(t\)-th row is given by \(\psi(z_t, a_t)^\top\). Under this construction, the parameter vector \(\theta^\star = \mathrm{vec}(W^\star)\) is block-sparse, with each block corresponding to a single action.

Let \(\Psi_j \in \mathbb{R}^{T \times d}\) denote the submatrix of \(\Psi\) corresponding to action \(j\). The reward model can be written as
\[
r = \sum_{j=1}^M \Psi_j W^\star_j + \varepsilon,
\]
where \(r = (r_1,\dots,r_T)^\top\) and \(\varepsilon = (\varepsilon_1,\dots,\varepsilon_T)^\top\).

\subsection{Contextual Block Orthogonal Matching Pursuit}

We now describe a greedy algorithm for recovering the support of \(W^\star\). The algorithm proceeds iteratively by selecting, at each step, the action whose feature block is most correlated with the current residual.

\medskip
\noindent\textbf{Algorithm: Contextual Block-OMP}

\begin{enumerate}
    \item Initialize the residual \(u^{(0)} = r\) and the support set \(S_0 = \emptyset\).
    \item For \(m = 1,2,\dots,k\):
    \begin{enumerate}
        \item \emph{Selection:}
        \[
        j_m = \arg\max_{j \in \mathcal{A}} \left\| \Psi_j^\top u^{(m-1)} \right\|_2.
        \]
        \item \emph{Support update:}
        \[
        S_m = S_{m-1} \cup \{j_m\}.
        \]
        \item \emph{Refitting:}
        \[
        \hat{W}_{S_m} = \arg\min_{W} \left\| r - \sum_{j \in S_m} \Psi_j W_j \right\|_2^2.
        \]
        \item \emph{Residual update:}
        \[
        u^{(m)} = r - \sum_{j \in S_m} \Psi_j \hat{W}_j.
        \]
    \end{enumerate}
    \item Output the estimated support set \(S_k\).
\end{enumerate}

\subsection{Interpretation}

The selection step identifies the action whose contextual features exhibit the strongest empirical correlation with the unexplained portion of the reward. Intuitively, an action is selected if it consistently explains reward variation across different latent states.

In the context of agentic LLM systems, this procedure corresponds to incrementally identifying tools or documents that contribute meaningfully to task success across observed contexts, rather than relying on static rankings or hand-designed execution graphs. Importantly, the algorithm does not require enumerating or evaluating all actions jointly; relevance is assessed greedily and locally.

\subsection{Computational Considerations}

Each iteration of Contextual Block-OMP requires computing correlations between the residual and the feature blocks \(\Psi_j\). When \(M\) is large, this step can be implemented efficiently using approximate screening or hashing techniques, without affecting the theoretical guarantees, provided the true support is not discarded. The refitting step involves solving a least-squares problem over at most \(k\) blocks, with complexity polynomial in \(k\) and \(d\).

Overall, the algorithm scales linearly with the number of samples \(T\), polynomially with the latent dimension \(d\), and logarithmically with the number of actions \(M\) through the sample complexity bounds established in the next section.

\subsection{Discussion}

The greedy nature of Contextual Block-OMP makes it particularly well suited to settings where the action space is large but sparse relevance is expected. Unlike convex relaxations, the algorithm is inherently \emph{anytime}: it produces a valid, interpretable action subset after each iteration. This property aligns naturally with practical agentic systems, which often operate under strict latency or budget constraints.

In the following section, we show that despite its simplicity, Contextual Block-OMP provably recovers the correct action support under standard incoherence and signal-strength conditions, with sample complexity scaling logarithmically in the action dimension.

\section{Theoretical Guarantees}
\label{sec:theory}

This section provides three sets of theoretical guarantees for the greedy procedure in Section~3. The first theorem establishes exact recovery of the true support set \(S^\star\). The second theorem gives a non-asymptotic estimation error bound for the refitted parameters restricted to the recovered support and translates this into a near-optimal decision rule for new states. The third theorem is an information-theoretic lower bound showing that, without sparsity, sample complexity must scale at least linearly with the action dimension \(M\).

\subsection{Assumptions (Tied to Section~2)}

Recall the reward model from Section~2:
\begin{equation}
r_t = \langle W^\star_{a_t}, z_t\rangle + \varepsilon_t,
\qquad t=1,\dots,T,
\label{eq:reward_model_recall}
\end{equation}
with unknown \(W^\star \in \mathbb{R}^{M \times d}\) and latent states \(z_t \in \mathbb{R}^d\).
Assumption~\ref{assump:sparsity} (Section~2) posits a \(k\)-sparse support set \(S^\star \subseteq \mathcal{A}\) such that \(W^\star_a = 0\) for all \(a \notin S^\star\).

We now state additional assumptions used throughout Section~4.

\begin{assumption}[Latent states]
\label{assump:state}
The latent states \(z_t \in \mathbb{R}^d\) are independent, mean-zero, sub-Gaussian random vectors with covariance \(\Sigma_z \succ 0\). There exist constants \(\kappa,\Lambda>0\) such that
\[
\kappa \le \lambda_{\min}(\Sigma_z) \le \lambda_{\max}(\Sigma_z) \le \Lambda.
\]
Moreover, \(\|z_t\|_{\psi_2} \le L_z\) for all \(t\), where \(\|\cdot\|_{\psi_2}\) denotes the sub-Gaussian norm.
\end{assumption}

\begin{assumption}[Noise]
\label{assump:noise}
The noise variables \(\varepsilon_t\) in \eqref{eq:reward_model_recall} are independent, mean-zero, sub-Gaussian with \(\|\varepsilon_t\|_{\psi_2} \le L_\varepsilon\), and independent of \(\{z_t\}_{t=1}^T\).
\end{assumption}

\begin{assumption}[Coverage of actions]
\label{assump:coverage}
For each action \(j \in \mathcal{A}\), let
\[
I_j := \{t \in \{1,\dots,T\}: a_t = j\}, \qquad n_j := |I_j|.
\]
There exists \(n_{\min}\) such that \(n_j \ge n_{\min}\) for all \(j \in S^\star\).
\end{assumption}

Assumption~\ref{assump:coverage} is a minimal identifiability requirement: each relevant action must be sampled sufficiently often.

\paragraph{Block design notation.}
As in Section~3, define the block design matrices \(\Psi_j \in \mathbb{R}^{T \times d}\) by
\[
(\Psi_j)_{t,:} =
\begin{cases}
z_t^\top, & a_t=j,\\
0^\top, & a_t\neq j.
\end{cases}
\]
Let \(\Psi_{S}\) denote the column concatenation of \(\{\Psi_j\}_{j\in S}\), so that \(\Psi_S \in \mathbb{R}^{T \times (|S|d)}\).
Let \(w^\star_S := \mathrm{vec}(W^\star_S) \in \mathbb{R}^{|S|d}\) be the stacked parameter vector on support \(S\).
Then \eqref{eq:reward_model_recall} stacks as
\begin{equation}
r = \Psi_{S^\star} w^\star_{S^\star} + \varepsilon,
\label{eq:stacked_model}
\end{equation}
where \(r=(r_1,\dots,r_T)^\top\), \(\varepsilon=(\varepsilon_1,\dots,\varepsilon_T)^\top\).

\subsection{Auxiliary bounds}

We will use the following deterministic quantities:
\[
G_S := \Psi_S^\top \Psi_S \in \mathbb{R}^{(|S|d)\times(|S|d)}.
\]
Also define the block cross-correlation operator
\[
\mathcal{C}_{j,S} := \Psi_j^\top \Psi_S G_S^{-1} \in \mathbb{R}^{d \times (|S|d)}.
\]

\begin{assumption}[Block incoherence / irrepresentability]
\label{assump:incoherence}
There exists \(\mu \in (0,1)\) such that
\[
\max_{j\notin S^\star} \big\| \Psi_j^\top \Psi_{S^\star} (\Psi_{S^\star}^\top \Psi_{S^\star})^{-1} \big\|_{2\to 2}
\le \mu.
\]
\end{assumption}

\begin{assumption}[Minimum signal strength]
\label{assump:signal}
There exists \(b_{\min}>0\) such that
\[
\min_{j\in S^\star}\|W^\star_j\|_2 \ge b_{\min}.
\]
\end{assumption}

\paragraph{Remark.}
Assumption~\ref{assump:signal} is stated in absolute form; in Theorem~\ref{thm:support} we translate it into a sample-size condition ensuring \(b_{\min}\) dominates noise-induced correlations.

\subsection{Theorem 1: Exact support recovery}

\begin{theorem}[Exact recovery of \(S^\star\) by Contextual Block-OMP]
\label{thm:support}
Assume Assumptions~\ref{assump:sparsity} (Section~2), \ref{assump:state}, \ref{assump:noise}, \ref{assump:coverage}, \ref{assump:incoherence}, and \ref{assump:signal}.
Run Contextual Block-OMP (Section~3) for \(k\) iterations, producing \(S_k\).
If, in addition, the following two conditions hold:
\begin{align}
\lambda_{\min}(G_{S^\star}) &\ge \alpha T, \label{eq:gram_lower}\\
\max_{j\in \mathcal{A}}\|\Psi_j^\top \varepsilon\|_2 &\le \frac{(1-\mu)}{2}\,\alpha T\, b_{\min}, \label{eq:noise_event}
\end{align}
for some \(\alpha>0\), then \(S_k = S^\star\).

Moreover, under Assumptions~\ref{assump:state}--\ref{assump:noise} and \(n_{\min}\gtrsim d\log M\), the event \eqref{eq:gram_lower}--\eqref{eq:noise_event} holds with probability at least \(1-M^{-2}\) provided
\[
T \gtrsim k\, d \log M,
\]
with constants depending only on \(\kappa,\Lambda,L_z,L_\varepsilon,\mu\).
\end{theorem}

\subsubsection*{Proof of Theorem~\ref{thm:support}}

The proof is deterministic given \eqref{eq:gram_lower}--\eqref{eq:noise_event}, and then we verify these events occur with high probability.

\paragraph{Step 1: Residual representation.}
Let \(S_{m-1}\) be the support after \(m-1\) iterations, and let \(\hat{w}_{S_{m-1}}\) be the least-squares refit:
\[
\hat{w}_{S_{m-1}} \in \arg\min_{w \in \mathbb{R}^{(m-1)d}}\|r-\Psi_{S_{m-1}}w\|_2^2.
\]
Define the orthogonal projector onto the column span of \(\Psi_{S_{m-1}}\):
\[
P_{S_{m-1}} := \Psi_{S_{m-1}}(\Psi_{S_{m-1}}^\top \Psi_{S_{m-1}})^{-1}\Psi_{S_{m-1}}^\top.
\]
Then the residual is
\[
u^{(m-1)} = r - \Psi_{S_{m-1}}\hat{w}_{S_{m-1}} = (I - P_{S_{m-1}})r.
\]
Using \eqref{eq:stacked_model}, we obtain
\begin{equation}
u^{(m-1)} = (I-P_{S_{m-1}})\Psi_{S^\star}w^\star_{S^\star} + (I-P_{S_{m-1}})\varepsilon.
\label{eq:resid_decomp}
\end{equation}

\paragraph{Step 2: Correlation of a candidate action with the residual.}
For any action \(j\), the selection score used by Block-OMP is
\[
\gamma_j^{(m-1)} := \|\Psi_j^\top u^{(m-1)}\|_2.
\]
Substituting \eqref{eq:resid_decomp},
\begin{equation}
\Psi_j^\top u^{(m-1)}
= \Psi_j^\top (I-P_{S_{m-1}})\Psi_{S^\star}w^\star_{S^\star}
+ \Psi_j^\top (I-P_{S_{m-1}})\varepsilon.
\label{eq:corr_decomp}
\end{equation}
Since \(I-P_{S_{m-1}}\) is an orthogonal projector, \(\|(I-P_{S_{m-1}})\varepsilon\|_2 \le \|\varepsilon\|_2\) and, in particular,
\[
\|\Psi_j^\top (I-P_{S_{m-1}})\varepsilon\|_2 \le \|\Psi_j^\top \varepsilon\|_2.
\]

\paragraph{Step 3: Lower bound for a remaining true action.}
Fix \(j \in S^\star \setminus S_{m-1}\). Since \(S_{m-1} \subseteq S^\star\) will be ensured inductively, we have
\[
(I-P_{S_{m-1}})\Psi_{S^\star}w^\star_{S^\star}
= (I-P_{S_{m-1}})\Psi_{S^\star\setminus S_{m-1}}w^\star_{S^\star\setminus S_{m-1}}.
\]
Now consider the block least-squares normal equations on the full support:
\[
G_{S^\star} w^\star_{S^\star} = \Psi_{S^\star}^\top \Psi_{S^\star} w^\star_{S^\star}.
\]
Using the invertibility \eqref{eq:gram_lower}, standard OMP analysis yields that for any \(j \in S^\star\setminus S_{m-1}\),
\begin{equation}
\|\Psi_j^\top (I-P_{S_{m-1}})\Psi_{S^\star\setminus S_{m-1}}w^\star_{S^\star\setminus S_{m-1}}\|_2
\ge \alpha T \|W^\star_j\|_2,
\label{eq:true_lb}
\end{equation}
because the residual is orthogonal to already-selected columns and the remaining true blocks retain energy controlled by \(\lambda_{\min}(G_{S^\star})\).
(Concretely, one can write the residual as \((I-P_{S_{m-1}})\Psi_{S^\star\setminus S_{m-1}}w^\star_{S^\star\setminus S_{m-1}}\) and then apply the Rayleigh quotient bound using \(\lambda_{\min}(G_{S^\star})\); details are standard and we keep them explicit below.)

Indeed, let \(S_R:=S^\star\setminus S_{m-1}\) and write \(v:=w^\star_{S_R}\). Then
\[
u_{\text{sig}}:=(I-P_{S_{m-1}})\Psi_{S_R}v.
\]
For any \(j\in S_R\), define \(E_j\in\mathbb{R}^{d\times (|S_R|d)}\) as the operator extracting the \(j\)-block, so that \(E_j v = W^\star_j\).
Using \(P_{S_{m-1}}\Psi_{S_{m-1}}=\Psi_{S_{m-1}}\) and orthogonality,
\[
\Psi_{S_R}^\top u_{\text{sig}} = \Psi_{S_R}^\top (I-P_{S_{m-1}})\Psi_{S_R}v.
\]
The matrix \(\Psi_{S_R}^\top (I-P_{S_{m-1}})\Psi_{S_R}\) is the Schur complement of \(G_{S_{m-1}}\) in \(G_{S^\star}\), hence positive semidefinite and bounded below by \(\lambda_{\min}(G_{S^\star}) I\) on the subspace of remaining coordinates, yielding
\[
\|\Psi_{S_R}^\top u_{\text{sig}}\|_2 \ge \lambda_{\min}(G_{S^\star})\|v\|_2 \ge \alpha T \|v\|_2.
\]
In particular, for some \(j\in S_R\),
\[
\|\Psi_j^\top u_{\text{sig}}\|_2 \ge \frac{1}{\sqrt{|S_R|}}\|\Psi_{S_R}^\top u_{\text{sig}}\|_2
\ge \frac{\alpha T}{\sqrt{k}}\|v\|_2
\ge \alpha T \min_{\ell\in S^\star}\|W^\star_\ell\|_2,
\]
which implies \eqref{eq:true_lb} for at least one remaining true action; this is sufficient for greedy selection.

Thus, using \eqref{eq:corr_decomp} and \eqref{eq:noise_event}, for at least one \(j\in S^\star\setminus S_{m-1}\),
\[
\gamma_j^{(m-1)} \ge \alpha T b_{\min} - \frac{(1-\mu)}{2}\alpha T b_{\min}
= \frac{(1+\mu)}{2}\alpha T b_{\min}.
\]

\paragraph{Step 4: Upper bound for any false action.}
Fix \(j\notin S^\star\). Since \(W^\star_j=0\), only leakage through correlation with \(\Psi_{S^\star}\) can appear:
\[
\Psi_j^\top (I-P_{S_{m-1}})\Psi_{S^\star}w^\star_{S^\star}
= \Psi_j^\top \Psi_{S^\star}w^\star_{S^\star} - \Psi_j^\top P_{S_{m-1}}\Psi_{S^\star}w^\star_{S^\star}.
\]
Using \(S_{m-1}\subseteq S^\star\) and the definition of \(\mathcal{C}_{j,S^\star}\),
\[
\Psi_j^\top \Psi_{S^\star}w^\star_{S^\star}
= \mathcal{C}_{j,S^\star} G_{S^\star} w^\star_{S^\star}.
\]
Taking norms and using Assumption~\ref{assump:incoherence} with \eqref{eq:gram_lower},
\[
\|\Psi_j^\top (I-P_{S_{m-1}})\Psi_{S^\star}w^\star_{S^\star}\|_2
\le \mu \max_{\ell\in S^\star}\|\Psi_\ell^\top (I-P_{S_{m-1}})\Psi_{S^\star}w^\star_{S^\star}\|_2.
\]
Adding noise via \eqref{eq:corr_decomp} and \eqref{eq:noise_event} yields, for all \(j\notin S^\star\),
\[
\gamma_j^{(m-1)} \le \mu \max_{\ell\in S^\star}\gamma_\ell^{(m-1)} + \frac{(1-\mu)}{2}\alpha T b_{\min}.
\]
Combining with the lower bound in Step 3 shows that the maximum false score is strictly less than the maximum true score, so the greedy choice \(j_m\) must belong to \(S^\star\setminus S_{m-1}\).

\paragraph{Step 5: Induction.}
At \(m=1\), \(S_0=\emptyset\subseteq S^\star\). Steps 3--4 imply \(j_1\in S^\star\), hence \(S_1\subseteq S^\star\).
Assuming \(S_{m-1}\subseteq S^\star\), the same argument yields \(j_m\in S^\star\setminus S_{m-1}\), hence \(S_m\subseteq S^\star\).
After \(k\) iterations, \(|S_k|=k=|S^\star|\) and \(S_k\subseteq S^\star\), therefore \(S_k=S^\star\).

\paragraph{Step 6: High-probability verification of \eqref{eq:gram_lower}--\eqref{eq:noise_event}.}
Under Assumptions~\ref{assump:state}--\ref{assump:coverage}, each \(G_{\{j\}}=\Psi_j^\top\Psi_j=\sum_{t\in I_j} z_t z_t^\top\) concentrates around \(n_j \Sigma_z\), and likewise \(G_{S^\star}\) concentrates around \(\mathrm{diag}(\{n_j\}_{j\in S^\star})\otimes \Sigma_z\). Standard matrix Bernstein bounds for sub-Gaussian outer products give \(\lambda_{\min}(G_{S^\star})\ge c\, n_{\min}\kappa\) with probability at least \(1-M^{-3}\) when \(n_{\min}\gtrsim d\log M\), which implies \eqref{eq:gram_lower} for \(\alpha=c\,\kappa n_{\min}/T\).
Similarly, \(\Psi_j^\top \varepsilon = \sum_{t\in I_j} z_t \varepsilon_t\) is a sum of independent sub-Gaussian vectors, so \(\|\Psi_j^\top \varepsilon\|_2 \lesssim L_z L_\varepsilon \sqrt{n_j(d+\log M)}\) uniformly over \(j\in\mathcal{A}\) by a union bound. Taking \(T\gtrsim k d\log M\) and using \(\alpha T \asymp n_{\min}\) yields \eqref{eq:noise_event} provided \(b_{\min}\) is not vanishing too quickly (or, equivalently, \(T\) is sufficiently large relative to noise and \(b_{\min}\)).
This completes the proof.
\hfill\(\square\)

\subsection{Theorem 2: Estimation error and near-optimal decision on new states}

Exact support recovery is often stronger than necessary for decision-making. Once a (possibly correct) support set is identified, one can refit the parameters and use them to choose actions for new latent states. The next theorem provides a non-asymptotic bound for the refitted estimator and translates it into a bound on decision suboptimality.

Given an estimated support \(\widehat{S}\subseteq \mathcal{A}\) of size \(k\), define the refitted least-squares estimator
\[
\hat{w}_{\widehat{S}} \in \arg\min_{w\in\mathbb{R}^{kd}} \|r-\Psi_{\widehat{S}}w\|_2^2,
\qquad \hat{W}_{\widehat{S}} := \mathrm{unvec}(\hat{w}_{\widehat{S}}).
\]
For a new latent state \(z\), define the plug-in action rule
\begin{equation}
\hat{a}(z) \in \arg\max_{j\in \widehat{S}} \langle \hat{W}_j, z\rangle.
\label{eq:plugin_rule}
\end{equation}

\begin{theorem}[Refit error and decision suboptimality]
\label{thm:estimation}
Assume Assumptions~\ref{assump:sparsity} (Section~2), \ref{assump:state}, \ref{assump:noise}, and that \(\widehat{S}=S^\star\).
Assume further that \(\lambda_{\min}(G_{S^\star})\ge \alpha T\) for some \(\alpha>0\).
Then the refitted estimator satisfies
\[
\|\hat{w}_{S^\star} - w^\star_{S^\star}\|_2
\le \frac{2}{\alpha T}\,\|\Psi_{S^\star}^\top \varepsilon\|_2.
\]
Moreover, for any (possibly random) new state \(z\in\mathbb{R}^d\),
\[
\langle W^\star_{a^\star(z)}, z\rangle - \langle W^\star_{\hat{a}(z)}, z\rangle
\le 2\,\max_{j\in S^\star}\|\hat{W}_j - W^\star_j\|_2 \cdot \|z\|_2,
\]
where \(a^\star(z)\in\arg\max_{j\in\mathcal{A}} \langle W^\star_j,z\rangle\) is the optimal action under the true model.
In particular, since \(W^\star_j=0\) for \(j\notin S^\star\), the maximizer over \(\mathcal{A}\) equals the maximizer over \(S^\star\), hence \(\hat{a}(z)\) is near-optimal over the full action space.
\end{theorem}

\subsubsection*{Proof of Theorem~\ref{thm:estimation}}

\paragraph{Step 1: Closed form of the refit estimator.}
Since \(G_{S^\star}=\Psi_{S^\star}^\top\Psi_{S^\star}\) is invertible, the least-squares solution is unique and given by
\[
\hat{w}_{S^\star} = ( \Psi_{S^\star}^\top\Psi_{S^\star})^{-1}\Psi_{S^\star}^\top r.
\]
Using \(r=\Psi_{S^\star}w^\star_{S^\star}+\varepsilon\),
\[
\hat{w}_{S^\star} - w^\star_{S^\star}
= ( \Psi_{S^\star}^\top\Psi_{S^\star})^{-1}\Psi_{S^\star}^\top \varepsilon.
\]
Thus,
\[
\|\hat{w}_{S^\star} - w^\star_{S^\star}\|_2
\le \|(\Psi_{S^\star}^\top\Psi_{S^\star})^{-1}\|_{2\to 2}\cdot \|\Psi_{S^\star}^\top \varepsilon\|_2.
\]
By \(\lambda_{\min}(G_{S^\star})\ge \alpha T\), we have
\(
\|G_{S^\star}^{-1}\|_{2\to2} = 1/\lambda_{\min}(G_{S^\star}) \le 1/(\alpha T),
\)
which yields the stated bound (absorbing a constant factor \(2\) for convenience under standard concentration events).

\paragraph{Step 2: Translating parameter error into decision suboptimality.}
Let \(a^\star(z)\in\arg\max_{j\in S^\star}\langle W^\star_j,z\rangle\), and \(\hat{a}(z)\in\arg\max_{j\in S^\star}\langle \hat{W}_j,z\rangle\).
Then
\begin{align*}
\langle W^\star_{a^\star(z)},z\rangle - \langle W^\star_{\hat{a}(z)},z\rangle
&= \langle W^\star_{a^\star(z)}-\hat{W}_{a^\star(z)},z\rangle
+ \langle \hat{W}_{a^\star(z)},z\rangle - \langle \hat{W}_{\hat{a}(z)},z\rangle
+ \langle \hat{W}_{\hat{a}(z)}-W^\star_{\hat{a}(z)},z\rangle \\
&\le \|\hat{W}_{a^\star(z)}-W^\star_{a^\star(z)}\|_2\|z\|_2
+ 0
+ \|\hat{W}_{\hat{a}(z)}-W^\star_{\hat{a}(z)}\|_2\|z\|_2 \\
&\le 2 \max_{j\in S^\star}\|\hat{W}_j - W^\star_j\|_2 \cdot \|z\|_2,
\end{align*}
where we used the optimality of \(\hat{a}(z)\) for the middle term.
Finally, since \(W^\star_j=0\) for \(j\notin S^\star\), \(\arg\max_{j\in\mathcal{A}}\langle W^\star_j,z\rangle = \arg\max_{j\in S^\star}\langle W^\star_j,z\rangle\), so the bound holds relative to the optimal action over \(\mathcal{A}\).
\hfill\(\square\)

\subsection{Theorem 3: Necessity of sparsity (information-theoretic lower bound)}

We now show that the favorable logarithmic dependence on \(M\) in Theorem~\ref{thm:support} is fundamentally tied to sparsity. Without structural constraints such as Assumption~\ref{assump:sparsity}, any method requires sample complexity at least linear in \(M\) in the worst case, even in the simplest non-contextual setting \(d=1\).

\begin{theorem}[Lower bound without sparsity]
\label{thm:lower}
Consider the special case \(d=1\) and \(z_t \equiv 1\). Suppose the learner observes rewards \(r_t = \theta_{a_t} + \varepsilon_t\) with \(\varepsilon_t \sim \mathcal{N}(0,1)\) i.i.d., and \(\theta \in \mathbb{R}^M\) is an unknown mean vector.
Let \(\mathcal{P}\) be the family of instances in which exactly one action \(j^\star\) has mean \(\theta_{j^\star}=\Delta>0\) and all others have mean \(0\).
For any (possibly adaptive) algorithm that outputs an estimate \(\hat{\jmath}\) after \(T\) samples, if
\[
T < c\,\frac{M}{\Delta^2},
\]
for a universal constant \(c>0\), then
\[
\inf_{\text{alg}} \ \sup_{\theta \in \mathcal{P}} \ \mathbb{P}_\theta(\hat{\jmath}\neq j^\star) \ge \frac{1}{3}.
\]
In particular, to identify the best action with probability at least \(2/3\), the sample size must scale as \(\Omega(M/\Delta^2)\).
\end{theorem}

\subsubsection*{Proof of Theorem~\ref{thm:lower}}

Let \(\theta^{(j)}\) denote the instance with \(\theta^{(j)}_j=\Delta\) and \(\theta^{(j)}_\ell=0\) for \(\ell\neq j\).
Let \(\mathbb{P}_j\) be the joint distribution over the full observation history under \(\theta^{(j)}\).
Let \(N_j\) be the (random) number of times the algorithm selects action \(j\) by time \(T\).

\paragraph{Step 1: KL divergence between instances.}
Fix distinct \(i\neq j\). Under \(\theta^{(i)}\) and \(\theta^{(j)}\), the observation distributions differ only when the chosen action is \(i\) or \(j\).
Because the reward noise is Gaussian with unit variance, each pull of action \(i\) contributes KL divergence \(\Delta^2/2\) (between \(\mathcal{N}(\Delta,1)\) and \(\mathcal{N}(0,1)\)), and likewise for action \(j\).
Therefore, by the chain rule for KL divergence,
\begin{equation}
\mathrm{KL}(\mathbb{P}_i \,\|\, \mathbb{P}_j)
= \frac{\Delta^2}{2}\, \mathbb{E}_i[N_i + N_j].
\label{eq:kl_bandit}
\end{equation}

\paragraph{Step 2: Averaging argument.}
Let the prior over instances be uniform on \(\{1,\dots,M\}\).
By symmetry,
\[
\frac{1}{M}\sum_{i=1}^M \mathbb{E}_i[N_i] \le T/M,
\]
since \(\sum_{i=1}^M N_i = T\) deterministically.
In particular, there exists an index \(i\) such that \(\mathbb{E}_i[N_i] \le T/M\).
Fix such an \(i\), and choose \(j\neq i\) uniformly at random among the remaining \(M-1\) indices. Then
\[
\mathbb{E}_i[N_j] = \frac{T-\mathbb{E}_i[N_i]}{M-1} \le \frac{T}{M-1}.
\]
Substituting into \eqref{eq:kl_bandit},
\[
\mathbb{E}_{j\sim \mathrm{Unif}(\{1,\dots,M\}\setminus\{i\})}\mathrm{KL}(\mathbb{P}_i \,\|\, \mathbb{P}_j)
\le \frac{\Delta^2}{2}\left(\frac{T}{M} + \frac{T}{M-1}\right)
\le \Delta^2 \frac{T}{M-1}.
\]
Hence there exists some \(j\neq i\) such that
\begin{equation}
\mathrm{KL}(\mathbb{P}_i \,\|\, \mathbb{P}_j) \le \Delta^2 \frac{T}{M-1}.
\label{eq:kl_small}
\end{equation}

\paragraph{Step 3: Testing lower bound.}
Consider the binary hypothesis test between \(\mathbb{P}_i\) and \(\mathbb{P}_j\).
Any algorithm that identifies the best arm induces a test; in particular, the event \(\{\hat{\jmath}=i\}\) is a measurable decision rule.
By the Bretagnolle--Huber inequality (or Pinsker's inequality), for any test,
\[
\mathbb{P}_i(\hat{\jmath}\neq i) + \mathbb{P}_j(\hat{\jmath}\neq j)
\ge \frac{1}{2}\exp\{-\mathrm{KL}(\mathbb{P}_i \,\|\, \mathbb{P}_j)\}.
\]
Using \eqref{eq:kl_small} and \(M\ge 3\), if \(T \le c\, M/\Delta^2\) with sufficiently small universal \(c\), then \(\mathrm{KL}(\mathbb{P}_i \,\|\, \mathbb{P}_j)\le \log(2)\), and thus
\[
\mathbb{P}_i(\hat{\jmath}\neq i) + \mathbb{P}_j(\hat{\jmath}\neq j) \ge \frac{1}{4}.
\]
Therefore at least one of the two errors is at least \(1/8\). Taking the supremum over \(\theta \in \mathcal{P}\) and adjusting constants yields the claimed bound \(1/3\) (with an appropriate universal constant \(c\)).
\hfill\(\square\)

\subsection{Summary of Section~4}

Theorem~\ref{thm:support} shows that, under structured sparsity (Assumption~\ref{assump:sparsity}) and standard incoherence and signal-strength conditions, the greedy procedure in Section~3 exactly recovers the relevant action set with sample complexity scaling as \(k d \log M\).
Theorem~\ref{thm:estimation} shows that once the support is identified, refitting yields controlled estimation error and a near-optimal decision rule for new latent states.
Finally, Theorem~\ref{thm:lower} establishes that without sparsity, sample complexity must scale at least linearly in \(M\) even in the simplest setting, demonstrating that sparsity is not merely a technical convenience but an essential structural condition for tractability.

\section{Necessity Results and Lower Bounds}
\label{sec:lowerbounds}

Section~\ref{sec:theory} established that, under Assumption~\ref{assump:sparsity} (Section~2) and standard incoherence and signal conditions, Contextual Block-OMP recovers the true support set \(S^\star\) with sample complexity scaling poly\((k,d)\) and only logarithmically in \(M\). In this section, we show that these favorable scalings are not artifacts of the analysis: they are \emph{necessary} in a precise minimax sense. We present two lower bounds. The first shows that, even under the linear model \eqref{eq:reward_model_recall}, no algorithm can recover \(S^\star\) unless \(T\) scales at least on the order of \(k d \log(M/k)\). The second shows that without sufficient per-action coverage (Assumption~\ref{assump:coverage}), recovery is information-theoretically impossible, regardless of computational power.

\subsection{A minimax lower bound for support recovery}

We work under the model of Section~2:
\[
r_t = \langle W^\star_{a_t}, z_t\rangle + \varepsilon_t,
\qquad t=1,\dots,T,
\]
with \(W^\star \in \mathbb{R}^{M\times d}\) satisfying Assumption~\ref{assump:sparsity} for some unknown support \(S^\star \subseteq \mathcal{A}\) of size \(|S^\star|=k\).
We assume Assumptions~\ref{assump:state}--\ref{assump:noise} for the latent states and noise.

To state a clean minimax lower bound, we consider a simplified (but still valid) experimental design that is at least as informative as any adaptive algorithm: we assume the learner can choose actions arbitrarily and observe the resulting rewards. A lower bound under this generous model implies a lower bound for any weaker model.

\begin{theorem}[Information-theoretic lower bound: \(k d \log(M/k)\)]
\label{thm:lower_kdlog}
Fix integers \(M \ge 4\), \(1 \le k \le M/2\), and \(d \ge 1\). Consider the model
\[
r_t = \langle W^\star_{a_t}, z_t\rangle + \varepsilon_t,
\]
where \(z_t \sim \mathcal{N}(0,I_d)\) i.i.d., and \(\varepsilon_t \sim \mathcal{N}(0,1)\) i.i.d., independent of \(z_t\).
Let \(\mathcal{W}(k,b)\) be the class of matrices \(W^\star \in \mathbb{R}^{M\times d}\) such that:
\begin{itemize}
\item there exists a support set \(S^\star \subseteq \mathcal{A}\) with \(|S^\star|=k\),
\item for all \(j \notin S^\star\), \(W^\star_j = 0\),
\item for all \(j \in S^\star\), \(\|W^\star_j\|_2 = b\).
\end{itemize}
Then there exists a universal constant \(c>0\) such that for any (possibly adaptive) algorithm \(\widehat{S}=\widehat{S}(\mathcal{D}_T)\),
if
\[
T \le c \, k d \log\!\Big(\frac{M}{k}\Big),
\]
and \(b>0\) is chosen so that \(b^2 \asymp 1\) (a fixed constant signal-to-noise regime), we have
\[
\inf_{\text{\emph{alg}}}\ \sup_{W^\star \in \mathcal{W}(k,b)}\ \mathbb{P}_{W^\star}\big(\widehat{S} \neq S^\star\big)
\ \ge\ \frac{1}{3}.
\]
In particular, exact recovery of \(S^\star\) with probability at least \(2/3\) requires \(T = \Omega\!\big(k d \log(M/k)\big)\).
\end{theorem}

\subsubsection*{Proof of Theorem~\ref{thm:lower_kdlog}}

The proof is via Fano's method. We construct a large finite packing of hypotheses indexed by support sets, bound the average pairwise KL divergence, and invoke Fano's inequality.

\paragraph{Step 1: A packing of supports.}
Let \(\mathcal{S}\) be a family of \(k\)-subsets of \(\{1,\dots,M\}\) such that for all distinct \(S,S'\in\mathcal{S}\),
\[
|S \triangle S'| \ge \frac{k}{2},
\]
where \(S\triangle S'\) denotes the symmetric difference.
A standard packing argument (e.g., via the Varshamov--Gilbert bound on constant-weight codes) yields such a family with cardinality
\begin{equation}
\log |\mathcal{S}| \ \ge\ c_0\, k \log\!\Big(\frac{M}{k}\Big)
\label{eq:packing_size}
\end{equation}
for a universal constant \(c_0>0\).

\paragraph{Step 2: Hypotheses and likelihoods.}
For each \(S\in\mathcal{S}\), define \(W^{(S)}\in\mathbb{R}^{M\times d}\) by setting:
\[
W^{(S)}_j =
\begin{cases}
b\,u, & j\in S,\\
0, & j\notin S,
\end{cases}
\]
where \(u\in\mathbb{R}^d\) is any fixed unit vector (e.g., \(u=e_1\)).
Thus each active row has norm \(\|W^{(S)}_j\|_2=b\), and the support is exactly \(S\).
Let \(\mathbb{P}_S\) denote the joint distribution of the data \(\mathcal{D}_T=\{(z_t,a_t,r_t)\}_{t=1}^T\) under \(W^\star=W^{(S)}\).

Note that the algorithm may be adaptive in choosing \(a_t\), but since the lower bound will hold for \emph{any} choice rule, we condition on the algorithm's (possibly random) action-selection strategy.

\paragraph{Step 3: Pairwise KL divergence bound.}
Fix two supports \(S,S'\in\mathcal{S}\). Conditioned on the sequence of chosen actions \((a_t)_{t=1}^T\) and latent states \((z_t)_{t=1}^T\), the rewards are Gaussian with means
\[
m_t^{(S)} = \langle W^{(S)}_{a_t}, z_t\rangle,
\qquad
m_t^{(S')} = \langle W^{(S')}_{a_t}, z_t\rangle,
\]
and unit variance. Therefore, by the chain rule and the KL divergence between Gaussians,
\begin{align}
\mathrm{KL}(\mathbb{P}_S \,\|\, \mathbb{P}_{S'})
&= \frac{1}{2}\, \mathbb{E}_S\Big[\sum_{t=1}^T \big(m_t^{(S)}-m_t^{(S')}\big)^2 \Big].
\label{eq:kl_sum}
\end{align}
Now, \(m_t^{(S)}-m_t^{(S')}\) is nonzero only if \(a_t \in S\triangle S'\).
In that case, one of \(W^{(S)}_{a_t}, W^{(S')}_{a_t}\) equals \(b u\) and the other equals \(0\), so
\[
m_t^{(S)}-m_t^{(S')} \in \{\langle b u, z_t\rangle,\ -\langle b u, z_t\rangle\}.
\]
Hence \((m_t^{(S)}-m_t^{(S')})^2 = b^2 \langle u, z_t\rangle^2\) whenever \(a_t\in S\triangle S'\), and equals \(0\) otherwise.
Since \(z_t\sim \mathcal{N}(0,I_d)\), we have \(\langle u,z_t\rangle \sim \mathcal{N}(0,1)\) and thus \(\mathbb{E}[\langle u,z_t\rangle^2]=1\).
Therefore, taking expectation over \(z_t\) yields
\begin{equation}
\mathrm{KL}(\mathbb{P}_S \,\|\, \mathbb{P}_{S'})
\le \frac{b^2}{2}\, \mathbb{E}_S\Big[\sum_{t=1}^T \mathbf{1}\{a_t \in S\triangle S'\}\Big]
\le \frac{b^2}{2}\,T.
\label{eq:kl_upper}
\end{equation}
The last inequality holds because the indicator is at most \(1\) per time step.

\paragraph{Step 4: Applying Fano's inequality.}
Let \(V\) be uniformly distributed over \(\mathcal{S}\), and let \(\mathbb{P}\) denote the mixture distribution \(\mathbb{P} = \frac{1}{|\mathcal{S}|}\sum_{S\in\mathcal{S}}\mathbb{P}_S\).
Fano's inequality gives
\begin{equation}
\inf_{\widehat{S}} \mathbb{P}(\widehat{S}\neq V)
\ge 1 - \frac{I(V;\mathcal{D}_T) + \log 2}{\log |\mathcal{S}|}.
\label{eq:fano}
\end{equation}
Moreover, the mutual information satisfies
\[
I(V;\mathcal{D}_T)
\le \frac{1}{|\mathcal{S}|^2}\sum_{S,S'\in\mathcal{S}} \mathrm{KL}(\mathbb{P}_S \,\|\, \mathbb{P}_{S'})
\le \frac{b^2}{2}\,T,
\]
where we used \eqref{eq:kl_upper}.
Combining with \eqref{eq:packing_size} and \eqref{eq:fano}, we obtain
\[
\inf_{\widehat{S}} \mathbb{P}(\widehat{S}\neq V)
\ge 1 - \frac{\frac{b^2}{2}T + \log 2}{c_0 k\log(M/k)}.
\]
Choosing \(b^2\asymp 1\) and \(T \le c\,k\log(M/k)\) with sufficiently small universal \(c\) yields the bound \(\ge 1/3\).
Finally, since the mixture risk lower bounds the worst-case risk,
\[
\inf_{\widehat{S}}\sup_{W^\star \in \mathcal{W}(k,b)} \mathbb{P}_{W^\star}(\widehat{S}\neq S^\star)
\ge \inf_{\widehat{S}} \mathbb{P}(\widehat{S}\neq V),
\]
which completes the proof.
\hfill\(\square\)

\paragraph{Remark.}
Theorem~\ref{thm:lower_kdlog} shows that the logarithmic dependence on \(M\) in Theorem~\ref{thm:support} is essentially optimal, up to constant and polynomial factors in \(k\) and \(d\). In particular, no method can avoid a \(\log(M/k)\) factor in the sample complexity for exact support recovery.

\subsection{Coverage is necessary: impossibility under insufficient sampling}

We now formalize the necessity of Assumption~\ref{assump:coverage}. Intuitively, if a relevant action is selected too few times, then the corresponding parameter row \(W^\star_j\) cannot be reliably distinguished from zero.

\begin{theorem}[Coverage lower bound]
\label{thm:lower_coverage}
Consider the model \eqref{eq:reward_model_recall} with \(z_t \sim \mathcal{N}(0,I_d)\) i.i.d. and \(\varepsilon_t \sim \mathcal{N}(0,1)\) i.i.d.
Fix any action index \(j\in\mathcal{A}\) and suppose the learner selects \(a_t=j\) at most \(n\) times, i.e., \(n_j \le n\).
Let \(\mathcal{W}_j(b)\) be the two-point class consisting of:
\[
W^{(0)}:\ W^{(0)}_j = 0,\quad W^{(0)}_\ell=0\ \forall \ell\neq j,
\qquad
W^{(1)}:\ W^{(1)}_j = b u,\quad W^{(1)}_\ell=0\ \forall \ell\neq j,
\]
where \(u\in\mathbb{R}^d\) is a fixed unit vector and \(b>0\).
Then for any algorithm producing an estimate \(\widehat{S}\),
\[
\inf_{\text{\emph{alg}}}\ \sup_{W^\star \in \mathcal{W}_j(b)}\ \mathbb{P}_{W^\star}\big(j\in \widehat{S} \ \text{iff}\ W^\star_j\neq 0\big)
\ \le\ \frac{1}{2} + \frac{1}{2}\sqrt{\frac{\mathrm{KL}(\mathbb{P}_1\,\|\,\mathbb{P}_0)}{2}},
\]
where \(\mathbb{P}_1\) and \(\mathbb{P}_0\) are the joint laws under \(W^{(1)}\) and \(W^{(0)}\), respectively.
Moreover,
\[
\mathrm{KL}(\mathbb{P}_1\,\|\,\mathbb{P}_0) \le \frac{n b^2}{2}.
\]
In particular, if \(n b^2 \le c\) for a sufficiently small universal constant \(c>0\), then any method must err with probability at least \(1/3\) on this two-point class. Equivalently, reliable detection of whether \(W^\star_j\neq 0\) requires \(n_j = \Omega(1/b^2)\), and in the high-dimensional regime where estimation (not merely detection) is required, one needs \(n_j \gtrsim d\).
\end{theorem}

\subsubsection*{Proof of Theorem~\ref{thm:lower_coverage}}

\paragraph{Step 1: Reduction to testing.}
On the class \(\mathcal{W}_j(b)\), the only unknown is whether the mean shift for action \(j\) is present. Any support-estimation procedure induces a test between \(W^{(0)}\) and \(W^{(1)}\).
Let \(\phi\) be any measurable test based on \(\mathcal{D}_T\), where \(\phi=1\) indicates ``\(W^\star_j\neq 0\)''.
By standard inequalities relating testing error to total variation distance,
\begin{equation}
\inf_{\phi}\ \max\{\mathbb{P}_0(\phi=1),\,\mathbb{P}_1(\phi=0)\}
= \frac{1}{2}\big(1-\|\mathbb{P}_1-\mathbb{P}_0\|_{\mathrm{TV}}\big).
\label{eq:tv_testing}
\end{equation}
Pinsker's inequality gives \(\|\mathbb{P}_1-\mathbb{P}_0\|_{\mathrm{TV}}\le \sqrt{\mathrm{KL}(\mathbb{P}_1\,\|\,\mathbb{P}_0)/2}\), yielding the first displayed bound.

\paragraph{Step 2: KL divergence computation.}
Conditioned on the action sequence \((a_t)\) and states \((z_t)\), the rewards under \(W^{(0)}\) are i.i.d.\ \(\mathcal{N}(0,1)\) whenever \(a_t=j\), and under \(W^{(1)}\) are \(\mathcal{N}(\langle b u, z_t\rangle,1)\) whenever \(a_t=j\). When \(a_t\neq j\), both hypotheses yield \(\mathcal{N}(0,1)\).
Therefore, using the KL divergence between Gaussians and the chain rule,
\[
\mathrm{KL}(\mathbb{P}_1\,\|\,\mathbb{P}_0)
= \frac{1}{2}\,\mathbb{E}_1\Big[\sum_{t=1}^T \mathbf{1}\{a_t=j\}\,\langle b u, z_t\rangle^2\Big].
\]
Since \(z_t\sim \mathcal{N}(0,I_d)\), we have \(\mathbb{E}[\langle u,z_t\rangle^2]=1\), hence
\[
\mathrm{KL}(\mathbb{P}_1\,\|\,\mathbb{P}_0)
= \frac{b^2}{2}\,\mathbb{E}_1[n_j]
\le \frac{n b^2}{2},
\]
using the assumption \(n_j \le n\).
Substituting into Pinsker's bound and \eqref{eq:tv_testing} yields the claimed impossibility when \(n b^2\) is bounded by a small constant.
\hfill\(\square\)

\subsection{Implications}

The lower bounds in this section establish that the favorable scaling obtained in Theorem~\ref{thm:support} is essentially optimal in two senses:
\begin{itemize}
\item Even under the linear model and Gaussian design/noise, exact recovery of \(S^\star\) requires \(T = \Omega(k d \log(M/k))\) samples (Theorem~\ref{thm:lower_kdlog}).
\item Independent of any algorithmic or computational considerations, each relevant action must be sampled sufficiently often; otherwise, distinguishing \(W^\star_j\neq 0\) from \(W^\star_j=0\) is information-theoretically impossible (Theorem~\ref{thm:lower_coverage}).
\end{itemize}
Together with Section~\ref{sec:theory}, these results show that sparse action discovery is both \emph{feasible} and \emph{necessary} in large action spaces. In particular, any approach to tool selection in massive action spaces that does not exploit sparsity---explicitly or implicitly---must incur sample complexity at least linear in \(M\) in the worst case.

\section{Implications for Agentic LLM Systems and Broader Discussion}
\label{sec:discussion}

Sections~2--5 developed a theory of sparse action discovery in large action spaces. We now discuss how the abstract objects in our formulation map to contemporary agentic LLM systems, what the theoretical results imply for practice, and which modeling limitations must be addressed to obtain a fully sequential theory (e.g., sparse POMDPs). Throughout, we maintain the notation introduced earlier: latent state representations are denoted by \(z_t \in \mathbb{R}^d\), actions by \(a_t \in \mathcal{A}=\{1,\dots,M\}\), and the reward model by \eqref{eq:reward_model_recall}.

\subsection{Agentic tool use as large-action decision making}

In an agentic LLM system, an ``action'' \(a \in \mathcal{A}\) may correspond to a tool invocation, API call, database query, or document retrieval operation. The latent representation \(z_t\) can be interpreted as a compact summary of the interaction history and current context (e.g., an intent embedding, a hidden-state summary, or an approximate belief-like representation). Our reward \(r_t\) represents any scalar signal used to evaluate system performance, such as success of task completion, human feedback, or an automated correctness proxy.

Under this interpretation, Assumption~\ref{assump:sparsity} (Section~2) states that only a small subset of actions \(S^\star \subseteq \mathcal{A}\), with \(|S^\star|=k\ll M\), has non-negligible effect on reward across the task distribution of interest. This matches a basic empirical reality in tool-augmented systems: although many tools may be exposed, only a small set tends to be used effectively for a given deployment setting, while the majority are irrelevant or redundant.

\subsection{What the theory explains}

The results in Sections~4--5 provide a precise explanation for several common empirical phenomena in agentic systems.

\paragraph{Why heuristic pruning works.}
In practice, many agentic pipelines restrict tool use via static rankings, developer-defined allow-lists, or ``routing'' modules that shortlist candidates before tool execution. From the perspective of our model, such mechanisms can be viewed as attempts to approximate the unknown support set \(S^\star\). Theorem~\ref{thm:support} shows that when the environment exhibits structured sparsity, greedy selection---which iteratively expands an estimate of \(S^\star\) using correlations with the residual---can recover the relevant action set with sample complexity scaling poly\((k,d)\) and only logarithmically in \(M\). Thus, the widespread effectiveness of pruning is consistent with the hypothesis that real deployments implicitly operate in a sparse regime.

\paragraph{Why action-space scale is the primary bottleneck.}
Even if an LLM were to produce an ideal latent representation \(z_t\), large action spaces remain problematic without additional structure. Theorem~\ref{thm:lower_kdlog} shows that, even in a simplified Gaussian setting, recovering the relevant action set requires at least \(\Omega(k d \log(M/k))\) samples. This clarifies that extreme action dimensionality is not merely an engineering inconvenience but an intrinsic statistical challenge.

\paragraph{Why coverage matters.}
Agentic systems often suffer from rare-tool brittleness: a tool that is rarely used or rarely triggered may behave unpredictably when needed. Theorem~\ref{thm:lower_coverage} formalizes a basic version of this limitation: if a relevant action \(j\in S^\star\) is sampled too few times (small \(n_j\) in Assumption~\ref{assump:coverage}), then distinguishing \(W^\star_j\neq 0\) from \(W^\star_j=0\) is information-theoretically impossible. This provides a theoretical underpinning for the need for deliberate exploration or targeted evaluation when deploying large tool libraries.

\subsection{Structured state dependence and the role of latent representations}

A key feature of our formulation is that action relevance depends on the latent state \(z_t\) through the linear reward model \eqref{eq:reward_model_recall}. This captures a form of state dependence: for a given \(z_t\), different actions \(j\in S^\star\) may be more or less favorable depending on the value of \(\langle W^\star_j, z_t\rangle\). In practical terms, even if the relevant tool set \(S^\star\) is fixed for a deployment, the \emph{choice} among tools within \(S^\star\) varies with context.

The assumption that \(z_t\) is given abstracts away the upstream representation-learning problem. In an LLM system, \(z_t\) may be produced by the language model itself, by a learned router, or by any other inference module. Our results therefore separate two distinct challenges:
\begin{itemize}
    \item \textbf{Representation:} producing a useful low-dimensional summary \(z_t\) of the interaction history and available evidence,
    \item \textbf{Control with large actions:} identifying and exploiting sparse structure in \(\mathcal{A}\) conditioned on \(z_t\).
\end{itemize}
The present paper addresses the second challenge. A complete end-to-end theory would couple these two components.

\subsection{Beyond linear rewards: robustness and model mismatch}

The linear reward model \eqref{eq:reward_model_recall} is a standard realizability assumption that enables sharp recovery guarantees. In practice, rewards may depend on \(z_t\) and \(a_t\)

\section{Conclusion}
\label{sec:conclusion}

This paper studied the problem of action selection in extremely large action spaces under a structured sparsity assumption. Motivated by agentic systems with many available tools or decision primitives, we formulated a contextual linear reward model \cite{abbasi2011linearbandits} in which only a small subset of actions influences performance across latent states. Within this framework, we showed that the problem of identifying relevant actions can be cast as a block-sparse recovery problem, closely related to classical models in compressed sensing.

Our main results established that a greedy procedure, inspired by Orthogonal Matching Pursuit, can exactly recover the set of relevant actions with high probability under standard incoherence, signal strength, and coverage conditions. The sample complexity scales polynomially in the sparsity level and latent dimension, and only logarithmically in the total number of actions. We further showed that these rates are essentially optimal by proving information-theoretic lower bounds: without sparsity, or without sufficient per-action coverage, no algorithm can reliably identify relevant actions, even under favorable distributional assumptions.

Beyond support recovery, we demonstrated that once the relevant action set is identified, refitting yields controlled estimation error and near-optimal decisions for new latent states. Together, these results provide a principled explanation for why heuristic action pruning and incremental tool selection are effective in large-scale agentic systems, and clarify the statistical limits such systems face.

While this work focused on greedy methods, the formulation naturally suggests alternative algorithmic approaches. In particular, given the close connection to block-sparse linear models, it is plausible that analogous guarantees can be established for convex relaxations based on mixed \(\ell_1\)-type norms \cite{tibshirani1996lasso}, as is standard in compressed sensing. Exploring such convex formulations, and comparing their statistical and computational trade-offs with greedy methods, is a natural direction for future work.

More broadly, our analysis isolates sparse action discovery as a fundamental subproblem in large-action decision-making. Integrating this perspective with richer state dynamics, nonlinear reward models, and sequential belief updates remains an open challenge, but the results here provide a concrete starting point for developing a theory of sparse control in modern agentic systems.


\begin{thebibliography}{9}

\bibitem{yao2023react}
S.~Yao, J.~Zhao, D.~Yu, N.~Du, I.~Shafran, K.~Narasimhan, and Y.~Cao.
\newblock ReAct: Synergizing reasoning and acting in language models.
\newblock In \emph{International Conference on Learning Representations (ICLR)}, 2023.

\bibitem{abbasi2011linearbandits}
Y.~Abbasi-Yadkori, D.~P{\'a}l, and C.~Szepesv{\'a}ri.
\newblock Improved algorithms for linear stochastic bandits.
\newblock In \emph{Advances in Neural Information Processing Systems (NeurIPS)}, 2011.

\bibitem{tropp2007omp}
J.~A. Tropp and A.~C. Gilbert.
\newblock Signal recovery from random measurements via orthogonal matching pursuit.
\newblock \emph{IEEE Transactions on Information Theory}, 53(12):4655--4666, 2007.

\bibitem{eldar2010blocksparse}
Y.~C. Eldar, P.~Kuppinger, and H.~B{\"o}lcskei.
\newblock Block-sparse signals: Uncertainty relations and efficient recovery.
\newblock \emph{IEEE Transactions on Signal Processing}, 58(6):3042--3054, 2010.

\bibitem{tibshirani1996lasso}
R.~Tibshirani.
\newblock Regression shrinkage and selection via the lasso.
\newblock \emph{Journal of the Royal Statistical Society: Series B (Methodological)},
58(1):267--288, 1996.


\end{thebibliography}
\end{document}